\documentclass{article} 
\usepackage[utf8]{inputenc}
\usepackage{iclr2026_conference,times}

\usepackage{amsmath,amsfonts,bm}

\def\eqref#1{equation~\ref{#1}}

\def\1{\bm{1}}

\DeclareMathAlphabet{\mathsfit}{\encodingdefault}{\sfdefault}{m}{sl}
\SetMathAlphabet{\mathsfit}{bold}{\encodingdefault}{\sfdefault}{bx}{n}

\usepackage{hyperref}
\usepackage{url}
\usepackage{framed}
\usepackage{tcolorbox}
\usepackage{svg}
\usepackage{algorithm}
\usepackage{algpseudocode}
\usepackage{wrapfig}
\usepackage{rotating}
\usepackage{amsmath}
\usepackage[font=small, skip=4pt]{caption}
\usepackage{enumitem}
\usepackage{amssymb}
\usepackage{multirow}
\usepackage{tcolorbox}
\tcbuselibrary{breakable, listingsutf8}  
\usepackage{listings}
\usepackage{adjustbox}
\title{From Experience to Strategy: Empowering LLM Agents with Trainable Graph Memory}

\author{Siyu Xia$^{1,2}$\thanks{Equal contribution.}, Zekun Xu$^{3}$\footnotemark[1],  Jiajun Chai$^{3}$, Wentian Fan$^{4}$, Yan Song$^{5}$, Xiaohan Wang$^{3}$\\ \textbf{ Guojun Yin$^{3}$ , Wei Lin$^{3}$, Haifeng Zhang$^{1,2}$\thanks{Corresponding author.Contact: haifeng.zhang@ia.ac.cn, jun.wang@cs.ucl.ac.uk},  Jun Wang$^{5}$\footnotemark[2]}\\
$^{1}$Institute of Automation, Chinese Academy of Sciences, Beijing, China \\
  $^{2}$School of Artificial Intelligence, University of Chinese Academy of Sciences, China\\
  $^{3}$Meituan \hspace{5pt} 
  $^{4}$Nanjing University of Posts and Telecommunications \hspace{2pt} \\
  $^{5}$AI Centre, Department of Computer Science, University College London, London, UK \hspace{2pt}
}

\iclrfinalcopy 
\begin{document}

\maketitle

\begin{abstract}
Large Language Models (LLMs) based agents have demonstrated remarkable potential in autonomous task-solving across complex, open-ended environments. A promising approach for improving the reasoning capabilities of LLM agents is to better utilize prior experiences in guiding current decisions. However, LLMs acquire experience either through implicit memory via training, which suffers from catastrophic forgetting and limited interpretability, or explicit memory via prompting, which lacks adaptability. In this paper, we introduce a novel \textbf{agent-centric, trainable, multi-layered graph memory} framework and evaluate how context memory enhances the ability of LLMs to utilize parametric information. The graph abstracts raw agent trajectories into structured decision paths in a state machine and further distills them into high-level, human-interpretable strategic meta-cognition. In order to make memory adaptable, we propose a reinforcement-based weight optimization procedure that estimates the empirical utility of each meta-cognition based on reward feedback from downstream tasks. These optimized strategies are then dynamically integrated into the LLM agent’s training loop through meta-cognitive prompting. Empirically, the learnable graph memory delivers robust generalization, improves LLM agents' strategic reasoning performance, and provides consistent benefits during Reinforcement Learning (RL) training.
\end{abstract}
\section{Introduction}

LLM-based agents are rapidly advancing the frontier of automated task execution, particularly in open-ended environments that demand long-horizon reasoning, strategic tool use, and adaptation from experience~\citep{yao2022react, gao2023toolformer,chai2025rlfactoryplugandplayreinforcementlearning}. While these agents demonstrate strong capabilities in decomposing and tackling complex tasks, their decision-making processes remain unstable, often resulting in inefficient action sequences, repeated mistakes, or even complete task failure ~\citep{singh2023progprompt}. A central challenge lies in empowering agents not only to act, but to continuously learn and adapt by extracting insights from past successes and errors.

Methods for enabling LLMs to better leverage prior experience can be broadly categorized into two paradigms. The first is \textbf{implicit memory}, is typically formed through training procedures like RL, which denotes LLMs encode syntactic structures and semantic relations into parameter space~\citep{li2025memosmemoryosai,bai2022traininghelpfulharmlessassistant}. A more intuitive alternative is \textbf{explicit memory} via contextual prompting, which improves performance by injecting guidance directly into the input without modifying model weights.~\citep{xu2025amemagenticmemoryllm,chhikara2025mem0buildingproductionreadyai,zhao2024expelllmagentsexperiential}. 

However, both paradigms suffer from fundamental yet contrasting limitations.
Explicit memory facilitates transparency by making reasoning steps externally visible through prompts; however, it lacks adaptability and struggles to generalize beyond specific tasks or contexts.
Conversely, implicit memory enables generalization via training, but its black-box nature makes the contribution of specific past experiences inaccessible and difficult to interpret, while encoding knowledge directly into parameter space often incurs information loss and is vulnerable to catastrophic forgetting.
This unresolved challenge motivates our central research question: \textit{Can we develop an agentic framework by leveraging dynamic, structured explicit memory to actively guide and enhance implicit policy learning?}

This paper introduces a novel \textbf{agent-centric, trainable, multi-layered graph memory} framework and explores its integration with RL. First, we abstract episodic agent trajectories into canonical paths over a finite state machine, from which we derive high-level, generalizable \textit{meta-cognition}. Second, we design a trainable graph architecture equipped with a reinforcement-driven weight optimization mechanism that calibrates the utility of stored strategies based on downstream task performance. Finally, the dynamic graph is operationalized as an \textit{explicit policy prior}, selectively injecting high-quality strategies into the agent's context during training. Empirical results across seven diverse question-answering benchmarks demonstrate that our framework delivers strong gains in both cross-task generalization and final task performance.

Our main contributions are threefold:
\begin{itemize}
        \item We propose a novel \textbf{agent-centric memory framework} that abstracts low-level agent trajectories into canonical paths on a finite state machine, enabling the distillation of \textbf{high-level, generalizable meta-cognitive} strategies.  
    
    \item We develop a \textbf{reinforcement-driven weight optimization mechanism} that dynamically calibrates the utility of memory connections, allowing the graph to selectively emphasize strategies with proven empirical effectiveness.  
    
    \item We demonstrate that incorporating this graph memory as an explicit policy prior within RL substantially enhances policy learning and final task performance.  

\end{itemize}

Ultimately, this work presents a unified framework for creating more adaptive, efficient, and strategically-aware agents that not only act, but learn and reason from a continually evolving repository of their own experiences.

\begin{figure}[t]
  \centering
  \includegraphics[width=\textwidth]{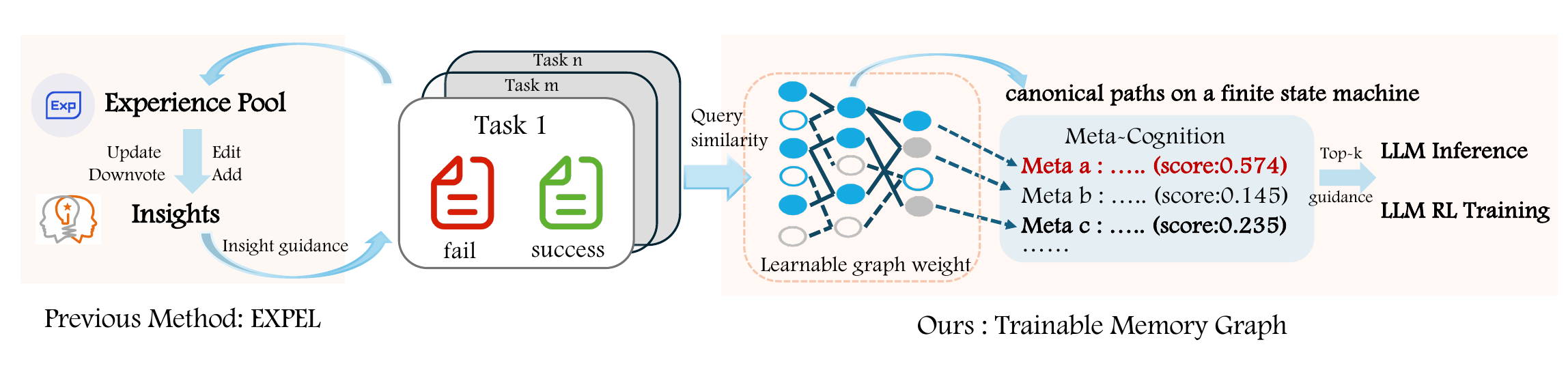}
  \caption{Our method and existing approach Expel~\citep{zhao2024expelllmagentsexperiential}.}
  \label{fig1:method comparison}
\end{figure}

\section{Related Work}
\label{sec:related_work}

\subsection{LLM Agents and Planning with External Tools}
LLM agents increasingly incorporate external tools to overcome reasoning limitations and expand their problem-solving capabilities. Early prompt-based approaches, including ReAct~\citep{yao2022react} and WebGPT~\citep{nakano2021webgpt}, demonstrate how agents can interleave reasoning and acting, embedding tool calls directly in the generation trace. Building on these foundations, Search-o1 introduces agentic RAG that dynamically retrieves knowledge during reasoning. Building on these foundations, Search-o1~\citep{li2025searcho1agenticsearchenhancedlarge} advances tool-augmented reasoning by enabling agents to autonomously decide when to invoke search tools during multi-step problem solving. Recent research has proposed more sophisticated coordination mechanisms using RL-based training~\citep{sun2025zerosearchincentivizesearchcapability,zheng2025deepresearcherscalingdeepresearch,song2025r1searcherincentivizingsearchcapability}. Search-R1~\citep{jin2025searchr1trainingllmsreason} represents a breakthrough RL framework that trains LLMs for alternating reasoning and search, enabling autonomous query generation and real-time information retrieval during step-by-step reasoning. Other recent approaches include optimized reward designs~\citep{wang2025actingreasoningmoreteaching, qian2025toolrlrewardtoollearning} and strategic tool integration~\citep{feng2025retoolreinforcementlearningstrategic}, with frameworks like RL-Factory~\citep{chai2025rlfactoryplugandplayreinforcementlearning} accelerating research in this domain.Despite these advances, the lack of explicit long-term memory for reusable tool-use patterns leaves deciding when and which tools to invoke as a key bottleneck. To address this limitation, we propose a differentiable graph-based memory system that encodes past decision paths into reusable strategic priors, enabling agents to systematically learn and generalize planning strategies across domains.

\subsection{Memory Architectures and Strategic Learning}

Recent research has increasingly explored how to extract strategic knowledge and meta-cognition from agent experience. Reflexion~\citep{zhang2023reflexion} equips agents with self-verbalized feedback to refine future behavior, while Expel~\citep{zhao2024expelllmagentsexperiential} identifies reusable reasoning trajectories to guide subsequent decisions. MEM1~\citep{zhou2025mem1learningsynergizememory} and MemAgent~\citep{yu2025memagentreshapinglongcontextllm} adapt memory usage over long-horizon tasks. A-MEM~\citep{xu2025amemagenticmemoryllm} builds dynamic memory notes that evolve with new inputs, Zep ~\citep{rasmussen2025zeptemporalknowledgegraph}and HopRAG~\citep{liu2025hopragmultihopreasoninglogicaware} construct logic-aware graphs to facilitate retrieval. 

However, these methods typically apply graph structure in a static manner and lack mechanisms to assess or refine the utility of memory components. G-Memory~\citep{zhang2025gmemorytracinghierarchicalmemory} demonstrates how hierarchical graph-based memory can evolve by assimilating new collaborative trajectories, enabling systems to leverage cross-trial knowledge and learn from prior experiences progressively.
~\cite{pan2025pastexperienceacceleratellm}focus on whether different forms of memory can enhance reasoning.~\citet{xiong2025memorymanagementimpactsllm} investigate long-term memory evolution.While prior memory methods often rely on static storage or task-specific designs, they lack mechanisms for evaluating and refining strategies.
In contrast, we propose a trainable graph-based memory that supports utility-aware strategy selection and reinforcement learning–driven updates, enabling generalizable and adaptive decision-making.

\section{Preliminaries}
\subsection{Heterogeneous Graph Structure}
Graphs provide a natural formalism for modeling structured dependencies among diverse entities. 
A heterogeneous graph ~\citep{zhang2019heterogeneous}can be defined as
\[
\mathcal{G} = (V, E, \mathcal{O}_V, \mathcal{R}_E, C),
\]
where $V$ denotes the set of nodes, $E \subseteq V \times V$ denotes the set of directed edges, $\mathcal{O}_V$ denotes the set of node types, $\mathcal{R}_E$ denotes the set of relation types, and $C$ is the collection of node contents. 
Each edge $e=(u,v,r) \in E$ specifies a relation of type $r$ from node $u$ to node $v$.  

Connectivity in $\mathcal{G}$ is represented by node-type adjacency matrices
\[
A^{xy} \in \{0,1\}^{|V_x|\times|V_y|}, \quad (x,y) \in \mathcal{O}_V \times \mathcal{O}_V,
\]
where $\mathcal{V}_x$ and $\mathcal{V}_y$ denote the sets of nodes of type $x$ and $y$, respectively. 
An entry $(A^{xy})_{ij}=1$ indicates that node $i$ of type $x$ is connected to node $j$ of type $y$.  
This formulation emphasizes the structural dependencies across different node types. 

To enable learning, each $A^{xy}$ is coupled with a weight matrix $W^{xy}$, so that propagation is governed by the weighted operator $A^{xy} \odot W^{xy}$. Thus, structure defines feasible paths, while weights determine effective information flow.
Formally,
\[
\mathbf{H}_{y} = \sigma\!\left((A^{xy} \odot W^{xy})^\top \mathbf{H}_{x}\right),
\]
where $\mathbf{H}_x$ are input values and $\sigma(\cdot)$ denotes an activation function.

\subsection{LLM Agents with Tool-Augmented Reasoning}

The interaction between a LLM and external tools can be formalized as a structured multi-turn decision process~\citep{chai2025rlfactoryplugandplayreinforcementlearning}.
At each time step $t$, the agent observes  
\[
s_t = (q, h_{1:t-1}), \quad 
a_t \sim \pi_\theta(a_t \mid s_t).
\]
where $q$ is the user query and $h_{1:t-1}$ is the dialogue or reasoning history, then generates an action $a$ which may correspond to internal reasoning, a tool invocation, or answer generation, using a protocol with tags such as \texttt{<think>}, \texttt{<tool\_call>}, and \texttt{<answer>}.

The process continues until either the tag \texttt{<answer></answer>} is generated or the agent has issued up to a maximum of $K$ tool invocations.  
A trajectory $\tau = (s_1,a_1,o_1,\dots,s_T,a_T,o_T)$ yields reward $R(\tau)$ , where $o_t$ denotes the environment observation, i.e., tool outputs if $a_t$ is a tool call, and the policy is optimized via  
\[
J(\theta) = \mathbb{E}_{\tau \sim \pi_\theta}[R(\tau)], \quad \nabla_\theta J(\theta) \approx \mathbb{E}_\tau\Big[\sum_{t=1}^T \nabla_\theta \log \pi_\theta(a_t \mid s_t)\,\hat{A}_t\Big].
\]
\section{Method}

\begin{figure}[t]
  \centering
  \includegraphics[width=\textwidth]{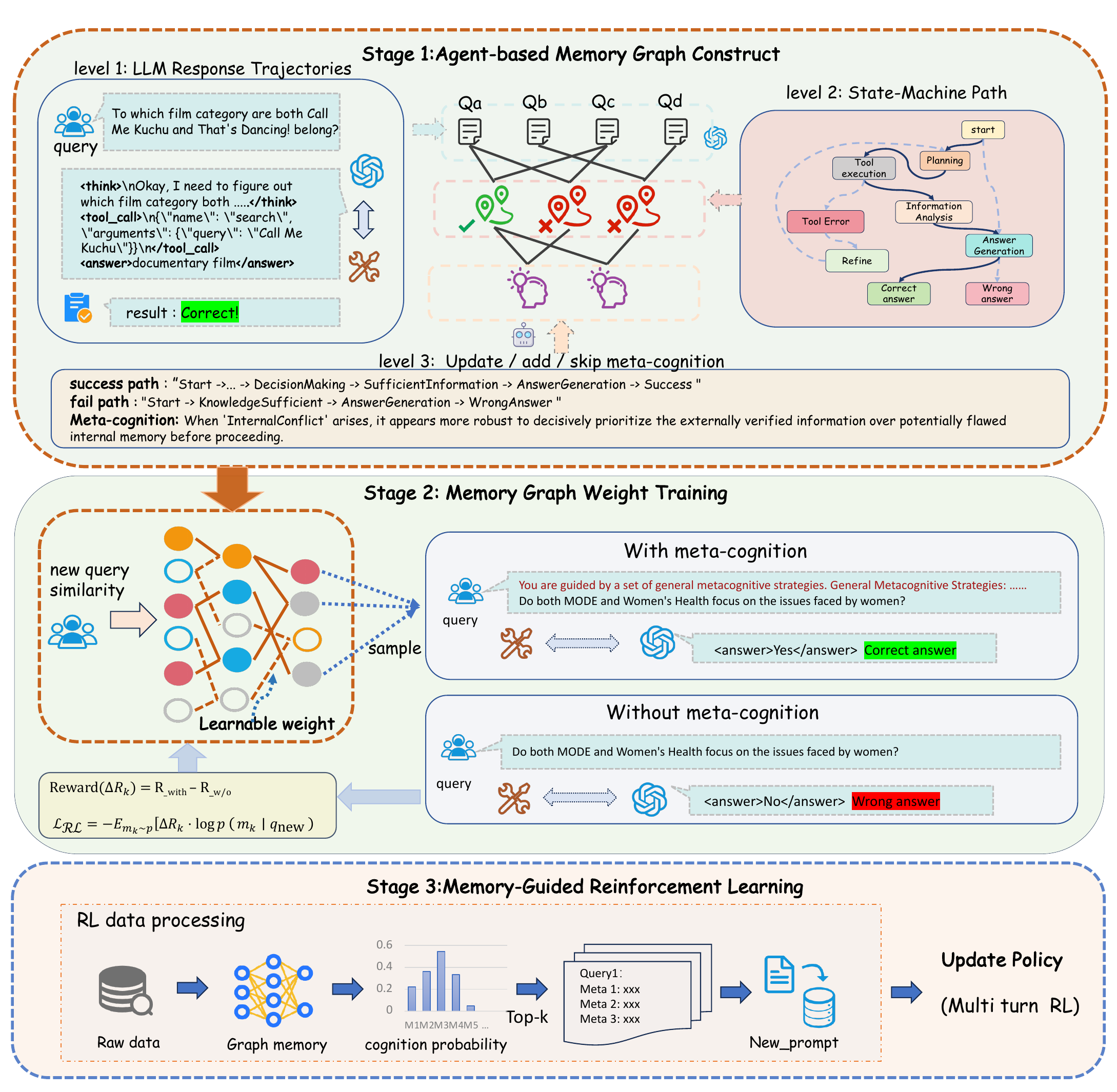}
  \caption{The framework of the proposed trainable memory.
Stage 1 builds a graph from LLM trajectories, encoding queries, decision paths, and meta-cognition.
Stage 2 estimates strategy utility via counterfactual rewards and updates graph weights.
Stage 3 injects top-k strategies into RL training for policy optimization.}
  \label{fig1:workflow of our method}
\end{figure}

In this section, we detail our proposed method in three stages. First, we describe how to construct a memory graph that encodes decision trajectories and strategic principles. Second, we present the learning framework for optimizing the weights within this memory graph. Finally, we explain how this structured memory is integrated into the RL training process to guide agent behavior and improve learning efficiency.The overall process of our method is shown in the Figure \ref{fig1:workflow of our method}.

\subsection{Stage 1: Hierarchical Memory Graph Construction}

\paragraph{Memory Graph Structure.} 
We instantiate the memory as a heterogeneous graph with node set 
$V = \mathcal{Q} \cup \mathcal{T} \cup \mathcal{M}$ 
and directed edges $E \subseteq (\mathcal{Q}\times\mathcal{T}) \cup (\mathcal{T}\times\mathcal{M})$.  
Each node type forms a distinct layer in the hierarchy, consistent with the structural depiction in figure \ref{fig1:workflow of our method} (stage 1):
\begin{itemize}
    \item \textbf{Query Layer} ($\mathcal{Q}$): Formed by query nodes $q_i$, each representing a task instance (e.g., a user query), including input, execution trajectory, and outcome labels.  
    Since a single query may yield multiple responses, each $q_i$ can connect to one or more transition paths $\{t_j\}$ via edges $(q_i \!\to\! t_j)$.  
    
    \item \textbf{Transition Path Layer} ($\mathcal{T}$): Formed by path nodes $t_j$, each denoting a canonical decision pathway derived from a finite state machine $\mathcal{S}$ that abstracts raw execution traces into standardized behavioral patterns.  
    These nodes are linked to the meta-cognition layer through edges $(t_j \!\to\! m_k)$.  
    
    \item \textbf{Meta-Cognition Layer} ($\mathcal{M}$): Formed by meta-cognition nodes $m_k$, each encoding a high-level strategic principle distilled from both successful and failed paths, serving as generalized heuristics for problem-solving.  
\end{itemize}

Connectivity is encoded by bipartite adjacency matrices
\[
A^{q\to t} \in \{0,1\}^{|\mathcal{Q}|\times|\mathcal{T}|}, \quad 
A^{t\to m} \in \{0,1\}^{|\mathcal{T}|\times|\mathcal{M}|},
\]
augmented with learnable weights $w_{qt}$ and $w_{tm}$ respectively.  
Information flows as a weighted aggregation process from queries to meta-cognition, forming a directed acyclic topology.

\paragraph{Finite State Machine.}  
To obtain a standardized and comparable representation of agent behaviors across tasks, we define a Finite State Machine (FSM) $\mathcal{S} = (S, A, T)$.  
Here, $S$ denotes abstract cognitive states (e.g., \texttt{StrategyPlanning}, \texttt{InformationAnalysis}), $A$ is the action space, and $T: S \times A \rightarrow S$ is the transition function.  
Each raw execution trajectory comprising tool invocations or reasoning steps is mapped onto a canonical path $t_j$ within this FSM.  
This grounding enables structured comparison across queries while filtering execution-level noise, ensuring that the memory graph preserves only semantically meaningful decision points.  
The detailed specification of $\mathcal{S}$ is provided in the Appendix~\ref{app:fsm}.

\paragraph{Meta-Cognition Induction.}
Meta-cognitions are induced by analyzing the canonical decision
pathways. For each query $q_i$, the agent samples trajectories $\{ \tau_1^{(i)}, \ldots, \tau_N^{(i)} \}$ from its policy $\pi$.  
If both successful ($\tau_s$) and failed ($\tau_f$) trajectories exist, contrasting their FSM paths yields a high-confidence meta-cognition $m_k$ that explains the outcome divergence.  
If only failures occur, the agent retrieves top-$K$ semantically similar queries 
$\text{Sim}(q_i,q_j)=\cos(\mathbf{e}_{q_i},\mathbf{e}_{q_j})$, 
and derives speculative meta-cognitions from successful paths of these neighbors:
\[
\mathcal{M}^{\text{spec}}(q_i) = \bigcup_{q_j \in \text{TopK}(q_i)} \{ m_k \mid t_j \in \text{SuccessPaths}(q_j),\, m_k \in \mathcal{M}(t_j) \}.
\]
Concrete examples and the corresponding prompts are provided in Appendix~\ref{prompt:meta-construction}.

\paragraph{Meta-Cognition Update.}
The memory graph is dynamically updated to preserve relevance and utility.  
When a new decision path is generated, the agent evaluates its strategic value: reinforcing existing principles updates their confidence, novel patterns lead to new meta-cognition nodes, and redundant or low-confidence paths are discarded.  
This selective process curates a concise set of strategic principles that evolves with experience.  

The hierarchical structure thus abstracts low-level trajectories into reusable strategies.  
At inference, the memory graph $\mathcal{G}$ functions as a structured policy prior guiding decision-making, while during training it provides supervision signals for reward-driven consolidation of meta-cognitive knowledge.

\subsection{Stage 2: Trainable Graph Weight Optimization}
\label{train-weight}

The memory graph provides structural priors, but not all meta-cognitions contribute equally. 
To adaptively capture their utility, we introduce a reinforcement-driven weight optimization procedure.

\paragraph{Parameterizing the Graph for Utility Learning.}
We parameterize the memory graph $\mathcal{G}$ as a sparsely connected weighted network, where each edge is associated with a trainable coefficient reflecting its utility.  
Given query features $\mathbf{H}_{\mathcal{Q}}^{(0)}$, information propagates through the graph via weighted aggregation:
\[
\mathbf{H}_{\mathcal{T}}^{(1)} = \sigma\!\left((A_{qt}\odot W_{qt})^\top \mathbf{H}_{\mathcal{Q}}^{(0)}\right), \quad
\mathbf{H}_{\mathcal{M}}^{(2)} = \sigma\!\left((A_{tm}\odot W_{tm})^\top \mathbf{H}_{\mathcal{T}}^{(1)}\right),
\]  
which corresponds to the flow from the \textit{query layer}, through the transition layer, and finally to the \textit{meta-cognition layer} in Figure~\ref{fig1:workflow of our method}.

In our formulation, a new query is represented by its similarity to historical queries in the graph, and the top-$k$ most relevant neighbors are selected to activate a task-specific subgraph $\mathcal{G}(q_{\text{new}}) = (\mathcal{Q}', \mathcal{T}', \mathcal{M}')$.Within this subgraph, a candidate meta-cognition $m_k \in \mathcal{M}'$ is sampled according to a relevance score $\rho(m_k)$, derived from the learned graph weights. 

To estimate its empirical utility, we contrast two trajectories: one guided by $m_k$, which yields reward $R_{\text{with}}(m_k)$, and another without such guidance, yielding reward $R_{\text{w/o}}$.  
The resulting \emph{reward gap}
$
\Delta R_k = R_{\text{with}}(m_k) - R_{\text{w/o}}
$
is employed as a utility signal, quantifying the marginal contribution of $m_k$ to overall task performance.

\paragraph{Policy Gradient-Based Weight Optimization.}
The relevance score $\rho(m_k \mid q_{\text{new}})$ is computed by aggregating path strengths from historical queries and transitions leading to $m_k$:
\[
\rho(m_k \mid q_{\text{new}}) = \sum_{q_i, t_j:\, q_i \to t_j \to m_k} \mathrm{Sim}(q_{\text{new}}, q_i) \cdot w_{qt}^{(i,j)} \cdot w_{tm}^{(j,k)}.
\]

Using a softmax over these scores, the selection probability$
p(m_k \mid q_{\text{new}}) \propto \exp(\rho(m_k \mid q_{\text{new}})).
$

We apply the REINFORCE algorithm to optimize the weights:
\[
\mathcal{L}_{\mathrm{RL}} = -\mathbb{E}_{m_k \sim p} \left[ \Delta R_k \cdot \log p(m_k \mid q_{\text{new}}) \right].
\]

A positive $\Delta R_k$ increases the relevance score and strengthens the supporting paths, while a negative $\Delta R_k$ decreases them, enabling the memory graph to refine itself over time.

\subsection{Stage 3: Memory-Guided Policy Optimization}

Departing from prior works that leverage memory solely during inference, our framework explicitly integrates the structured memory into the \textit{training loop}. Meta-cognitive strategies are dynamically retrieved from the optimized memory graph and incorporated into the agent's context, serving as high-level strategic priors that guide the reinforcement learning process.

\paragraph{Strategic Context Retrieval.}
For each training instance $q_{\text{train}}$, we compute a relevance score for every meta-cognition node $m \in \mathcal{M}$. This score is derived from the aggregated weights of all paths connecting the corresponding query node to the meta-cognition node within the memory graph $\mathcal{G}$ (as formulated in Section~\ref{train-weight}). We then select the top-$k$ meta-cognitions $\{m_1, \dots, m_k\}$ with the highest scores. This mechanism ensures that the guidance is not only relevant but also grounded in empirically successful past trajectories, as encoded by the learned edge weights.

The retrieved strategies are verbalized and prepended to the original query to form an augmented prompt, $\tilde{q}_{\text{train}}= \big[\, m_1, m_2, \dots, m_k \, ; \; q_{\text{train}} \,\big]$, this augmented prompt serves as the input to the policy network.

\paragraph{Optimization Objective.}
The agent's policy, $\pi_{\theta}$, is optimized to maximize the expected cumulative reward conditioned on the augmented context. We employ a policy gradient method, where the parameters $\theta$ are updated by minimizing the following loss function:
\[
\mathcal{L}_{\text{RL+Mem}} = - \mathbb{E}_{a \sim \pi_{\theta}(\cdot \mid \tilde{q}_{\text{train}})} \big[ R(a) \big].
\]
This tight integration ensures that the policy does not learn in isolation but is continually guided by a dynamically evolving corpus of strategic knowledge. This allows the agent to effectively bootstrap its learning process from a distilled representation of past successes.

In practice, we adopt the Generalized Reinforcement Policy Optimization (GRPO) algorithm to optimize the memory-augmented policy. 
The GRPO loss can be written as:
\[
\mathcal{L}_{\text{GRPO}} = - \mathbb{E}_{t} \left[
    \min\!\left(
        \frac{\pi_{\theta}(a_t \mid \tilde{q}_{\text{train}})}{\pi_{\theta_{\text{old}}}(a_t \mid \tilde{q}_{\text{train}})} \hat{A}_t,\;
        \text{clip}\!\left(\frac{\pi_{\theta}(a_t \mid \tilde{q}_{\text{train}})}{\pi_{\theta_{\text{old}}}(a_t \mid \tilde{q}_{\text{train}})}, 1-\epsilon, 1+\epsilon\right)\hat{A}_t
    \right)
\right],
\]
where $\hat{A}_t$ is the advantage estimator and $\epsilon$ the clipping parameter.

\section {Experiment}

\subsection{Datasets}
To evaluate the effectiveness and generalizability of our approach, we conduct experiments on seven widely-used Question-Answering(QA) datasets, covering both single-turn and multi-hop reasoning tasks.
\textbf{(1) General QA Datasets}: We include Natural Questions~\citep{kwiatkowski2019natural}, 
TriviaQA~\citep{joshi2017triviaqa}, and PopQA~\citep{mallen2022popqa}, which consist of open-domain factoid questions requiring retrieval and basic reasoning capabilities.
\textbf{(2) Multi-hop QA Datasets}: For more complex reasoning scenarios, we adopt HotpotQA~\citep{yang2018hotpotqa}, 2WikiMultiHopQA~\citep{ho2020constructing}, Musique~\citep{trivedi2022musique}, and Bamboogle~\citep{press2022measuring}, which require integrating information across multiple documents.

\renewcommand{\arraystretch}{1.3} 
\begin{table*}[t]
\centering
\caption{Performance comparison across seven QA datasets in inference. $^\dagger$ indicates in-domain datasets, while $^\star$ denotes out-of-domain datasets. Percentages in Avg. column denote relative improvement over ITR.}
\label{tab:inference-results}
\resizebox{\textwidth}{!}{
\begin{tabular}{l|cccccccc}
\hline
\multirow{2}{*}{Methods} & \multirow{2}{*}{Avg. (↑ / ↓ vs. ITR)} & \multicolumn{3}{c|}{General QA} & \multicolumn{4}{c}{Multi-Hop QA} \\ \cline{3-9}
 &  & NQ$^\star$ & TriviaQA$^\star$ & PopQA$^\star$ & HotpotQA$^\dagger$ & 2wiki$^\star$ & Musique$^\star$ & Bamboogle$^\star$  \\ \hline
\multicolumn{9}{l}{\textbf{Qwen3-8B}} \\
ITR               & 0.334 (–) &   0.275  &  0.593  &   0.358  &   0.325  &   0.324  &  0.094  &  0.365  \\
Direct Inference  & 0.269 (↓19.5\%) & 0.200 & 0.519 & 0.191 & 0.230 & 0.275 & 0.058 & \textbf{0.410} \\
CoT               & 0.252 (↓24.6\%) & 0.209 & 0.512 & 0.182 & 0.223 & 0.271 & 0.055 & 0.308 \\
Direct Trajectory & \underline{0.352 (↑5.4\%)} & \textbf{0.317} & \underline{0.604} & \underline{0.380} & 0.329 & \textbf{0.363} & 0.105 & 0.364 \\
A-MEM             & 0.334 (0.0\%) & 0.286 & 0.590 & 0.366 & \underline{0.339} & 0.332 & \underline{0.112} & 0.313 \\
EXPEL             & 0.329 (↓1.5\%) & 0.306 & 0.594 & 0.379 & 0.317 & 0.327 & 0.092 & 0.287 \\
Ours              & \textbf{0.365(↑9.3\%)} & \underline{0.316} & \textbf{0.622} & \textbf{0.382} & \textbf{0.358} & \underline{0.354} & \textbf{0.128} & \underline{0.392} \\ \hline

\multicolumn{9}{l}{\textbf{Qwen3-4B}} \\
ITR               & 0.279 (–) & 0.298 & 0.581 & 0.157 & 0.268 & 0.281 & 0.077 & 0.290 \\
Direct Inference  & 0.211 (↓24.4\%) & 0.158 & 0.413 & 0.157 & 0.183 & 0.240 & 0.033 & 0.290 \\
CoT               & 0.181 (↓35.1\%) & 0.149 & 0.375 & 0.146 & 0.156 & 0.190 & 0.022 & 0.228 \\
Direct Trajectory & \underline{0.325 (↑16.5\%) }& 0.310 & 0.558 & 0.379 & 0.282 & 0.344 & 0.076 & \underline{0.327} \\
A-MEM             & 0.319 (↑14.3\%) & 0.310 & \underline{0.586} & 0.381 & 0.272 & 0.269 & \underline{0.091} & 0.325 \\
EXPEL             & 0.321 (↑15.1\%) & \underline{0.312} & 0.570 & \underline{0.388} & \underline{0.294} & \textbf{0.347} & 0.075 & 0.263 \\
Ours              & \textbf{0.351 (↑25.8\%)} & \textbf{0.335} & \textbf{0.596} & \textbf{0.393} & \textbf{0.299} & \textbf{0.347} & \textbf{0.099} & \textbf{0.391} \\ \hline
\end{tabular}
}
\end{table*}

\subsection{Baseline Evaluation}
To comprehensively evaluate the effectiveness of our proposed method, we design experiments from two complementary perspectives:  
\textbf{(1) Direct Inference Impact:} We assess how the integration of our memory workflow influences model performance in zero-training settings, i.e., during direct inference.  
\textbf{(2) Training Impact:} We investigate how the memory architecture affects RL training dynamics, focusing on convergence speed and the final performance achieved. Detailed baseline configurations are provided in Appendix \ref{set}.

\subsection{Main results}

\textbf{Experimental Analysis: Memory-Guided Inference}. The detailed inference results are summarized in Table~\ref{tab:inference-results}. On the 8B-scale model, our method demonstrates strong competitiveness, achieving an average score of 0.365, which represents a notable \textbf{+9.3\%} relative improvement over the ITR baseline and ranks first among all contenders.
The advantages of our method become even more dramatic on the smaller Qwen3-4B model. It achieves a staggering \textbf{+25.8\%} relative improvement in average performance over the ITR baseline, this significant performance improvement on a model with limited capacity suggests that our method effectively addresses its inherent deficiencies by providing a robust and structured reasoning framework.

A particularly noteworthy finding is that the memory component of our method was constructed exclusively using data from HotpotQA, the single in-domain dataset. Despite this, our method not only excels on HotpotQA but also achieves state-of-the-art or highly competitive performance across all out-of-domain datasets, including NQ, TriviaQA, PopQA, and 2wiki. This outcome is a strong testament to the remarkable generalization capability of our approach. It demonstrates that the reasoning structures learned from HotpotQA are not merely overfitted patterns.

\renewcommand{\arraystretch}{1.3} 
\begin{table*}[htbp]
\centering
\caption{Performance comparison across seven QA datasets in training. 
Avg. column also reports relative improvement (\%) compared to Search-R1 as the base. 
$^\dagger$ indicates in-domain datasets, while $^\star$ denotes out-of-domain datasets.}
\label{tab:training-results}
\resizebox{\textwidth}{!}{
\begin{tabular}{l|c|ccc|cccc}
\hline
\multirow{2}{*}{Methods} & \multirow{2}{*}{Avg. (↑ / ↓ vs. Search-R1)} & \multicolumn{3}{c|}{General QA} & \multicolumn{4}{c}{Multi-Hop QA} \\ \cline{3-9}
 & & NQ$^\star$ & TriviaQA$^\star$ & PopQA$^\star$ & HotpotQA$^\dagger$ & 2wiki$^\star$ & Musique$^\star$ & Bamboogle$^\star$ \\ \hline
\multicolumn{9}{l}{\textbf{Qwen3-8B}} \\ 
Search-R1         & 0.395 (–)        & 0.384 & 0.651 & 0.429 & \textbf{0.391} & 0.386 & \underline{0.143} & 0.380 \\
Direct Trajectory & 0.400 (↑1.27\%)  & \textbf{0.406} & \underline{0.657} & 0.433 & 0.376 & 0.367 & 0.139 & \underline{0.423} \\
A-MEM             & \underline{0.403(↑2.03\%) }& \underline{0.398} & 0.656 & \textbf{0.436} & \underline{0.389} & \textbf{0.409} & 0.138 & 0.398 \\
EXPEL             & 0.371 (↓6.08\%)  & 0.362 & 0.621 & 0.407 & 0.354 & 0.375 & 0.121 & 0.357 \\
Ours              & \textbf{0.408 (↑3.29\%)}  & 0.386 & \textbf{0.662} & \underline{0.434} & 0.387 & \underline{0.403} & \textbf{0.152} & \textbf{0.435} \\ \hline

\multicolumn{9}{l}{\textbf{Qwen3-4B}} \\
Search-R1         & 0.375 (–)        & 0.357 & \underline{0.625} & 0.426 & 0.354 & 0.402 & 0.115 & 0.348 \\
Direct Trajectory & \underline{0.415 (↑10.67\%)} & 0.403 & 0.624 & 0.434 & \textbf{0.420} & \textbf{0.428} & \underline{0.186} & 0.412 \\
A-MEM             & 0.388 (↑3.47\%)  & \underline{0.393} & 0.603 & \underline{0.439} & 0.385 & 0.322 & 0.157 & \underline{0.418} \\
EXPEL             & 0.337 (↓10.13\%) & 0.322 & 0.577 & 0.399 & 0.311 & 0.363 & 0.081 & 0.305 \\
Ours              & \textbf{0.426 (↑13.60\%)} & \textbf{0.408} & \textbf{0.646} & \textbf{0.462} & \underline{0.410} & \underline{0.407} & \textbf{0.189} & \textbf{0.463} \\ \hline
\end{tabular}
}
\end{table*}

\textbf{Experimental Analysis: Memory-Guided Reinforcement Learning}. We further evaluated our method by integrating it into RL training process. The detailed training results are in Table~\ref{tab:training-results}.

On the \texttt{Qwen3-8B} model, our method achieves the best average performance (0.408), improving upon \texttt{Search-R1} baseline by \textbf{3.29\%}. This shows that our method provides additional benefits even after the model is already optimized with RL. The gains are most notable on challenging out-of-domain datasets like \texttt{TriviaQA} and \texttt{Bamboogle}, suggesting our memory helps the RL agent learn more general reasoning strategies that transfer well to new tasks.

\begin{wrapfigure}{r}{0.5\textwidth}  
  \centering
   \captionsetup{margin=2em} 
  \vspace{-10pt}  
  \includegraphics[width=\linewidth]{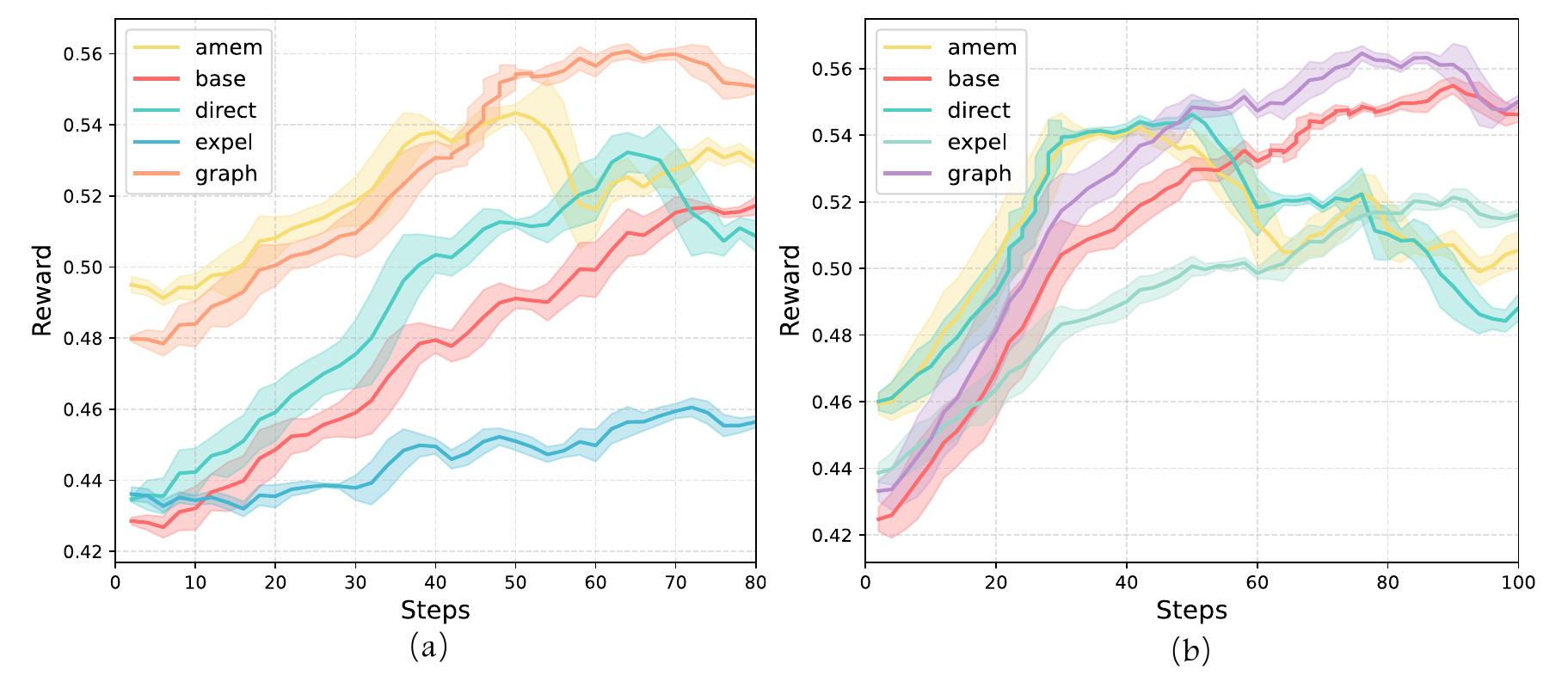}
 \caption{
  \small 
  (a) Training curve of 4B models. \\
  \hspace{1em} 
  (b) Training curve of 8B models.
}

  \label{fig:ablation-training-curve}
  \vspace{-10pt}  
\end{wrapfigure}

On the smaller \texttt{Qwen3-4B} model, the results are even more impressive. Our method achieves a remarkable \textbf{13.60\%} relative improvement over \texttt{Search-R1}. As seen in our inference experiments, the benefit of our method is especially pronounced on smaller models. Remarkably, our trained \texttt{Qwen3-4B} model (0.426) outperforms the baseline \texttt{Qwen3-8B} model (0.395), demonstrating a significant gain in efficiency.

In summary, adding our method to inference or RL training framework significantly boosts QA performance, especially for smaller models. Our structured memory helps the model learn general reasoning skills from the in-domain \texttt{HotpotQA} data and apply them successfully to other datasets. This allows smaller models to match or even exceed the performance of larger ones, offering a path to more efficient and capable models.

\subsection{Ablation Studies}

We conduct ablation studies across three dimensions: (1) disabling memory weight updates (2) varying the number of meta-cognitions used as context and (3) altering the granularity of memory composition (i.e., API call structure).

\begin{figure}[htbp]
  \centering
  \includegraphics[width=\textwidth]{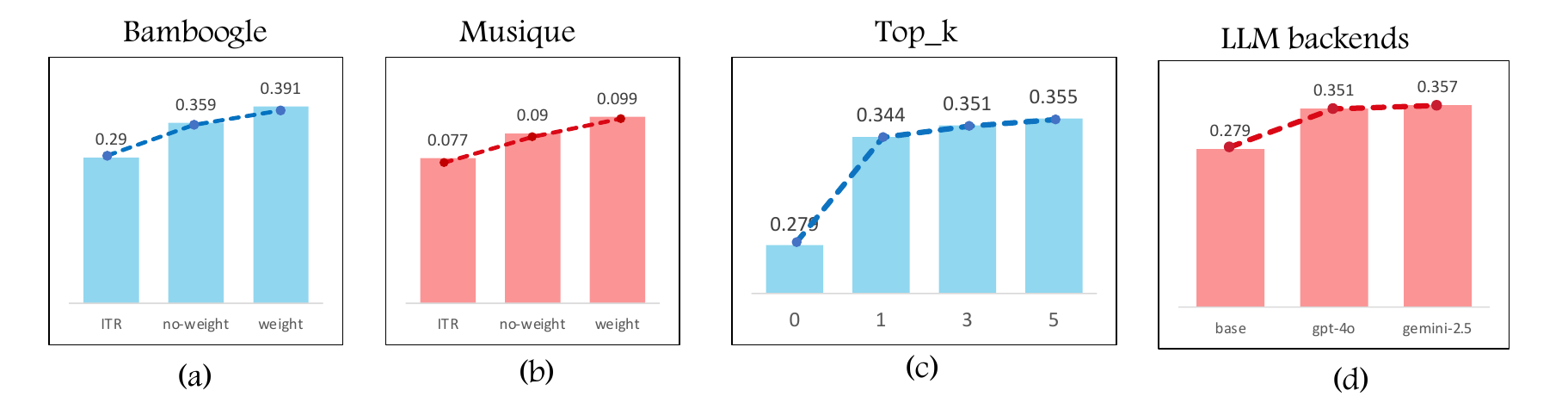}
  \caption{\textbf{Ablation studies of the structured memory framework.} 
 (a) and (b) show the effect of disabling weight optimization. (c) varying the number of meta-cognition $k$. (d) generalization across LLM backends.
  }
  \label{fig1:env}
\end{figure}

\textbf{Effect of Disabling Weight Optimization.}
We first examine the impact of freezing the memory graph weights (i.e., no learning of edge confidence). In this setup, we keep all memory edges at uniform weight and retrieve strategies purely based on structural presence. As shown in figure ~\ref{fig1:env}(a)(b), performance drops significantly, particularly on \texttt{2WikiMultiHopQA}, indicating that learning to prioritize high-utility memory connections is crucial for effective strategy reuse. This validates our reinforcement-based update mechanism, which helps distinguish broadly useful meta-cognitions from less effective or overly specific ones.

\textbf{Varying the Number of Meta-Cognitions.}
We further evaluate how the number of retrieved meta-cognitive strategies ($k$) affects model performance. Figure ~\ref{fig1:env}(c) presents the average accuracy of the 4B model on seven benchmarks as a function of the number of meta-cognitions. Increasing $k$ from 0 (no memory) to 3 leads to steady improvement, as more strategic signals are injected into the prompt. However, further increasing $k$ yields diminishing returns and can even introduce noise due to overlapping or irrelevant strategies. This highlights a trade-off between strategy diversity and clarity, and suggests that a moderate value of $k=3$ offers the best balance between guidance and prompt efficiency. The detailed results are shown in Table~\ref{tab:meta-number-ablation}.

\textbf{Generalization across LLMs backends.}
To evaluate whether our memory construction is tied to a specific LLM API, we replace the original \texttt{OpenAI gpt-4o} model with \texttt{Gemini-2.5-pro} and rerun the downstream evaluation using the same memory graph. As shown in Table~\ref{tab:diff-api}, our memory-augmented approach consistently outperforms its non-memory counterpart even under a different LLM backend, though the absolute numbers differ slightly due to model capability gaps. This demonstrates that our structured memory graph and retrieval-guided prompting strategy are largely \textit{model-agnostic}, enabling plug-and-play use across modern foundation models.

\section{Conclusion}

In this paper, we address the dual challenges of inefficient decision-making and poor experience reuse in LLM-based agents. We introduce a trainable, multi-level graph memory framework that structurally encodes historical queries, policy trajectories, and high-level metacognitive strategies. This design facilitates explicit strategy recall and integrates memory into the RL loop to guide and accelerate policy optimization.

Unlike prior works that rely on either implicit optimization or static prompting, our approach unifies explicit memory with dynamic learning. By updating memory weights via RL signals, the framework selectively reinforces high-utility strategies and re-injects them into the agent's training process through prompt augmentation. This mechanism promotes strategic transfer and generalization from past experiences. Our experiments demonstrate that this method not only improves reasoning accuracy at inference time but also accelerates convergence during RL training, ultimately yielding superior final performance and strong generalization across diverse tasks.
\newpage
\bibliography{iclr2026_conference}
\bibliographystyle{iclr2026_conference}

\appendix


\section{Experimental Setup Details}

\subsection{multi-turn tool-intergrated QA}
When tackling QA benchmarks, we observe that incorporating external knowledge retrieval plays a crucial role in enhancing answer accuracy. To this end, we utilize the 2018 Wikipedia dump as our external knowledge base and adopt the E5 retriever for efficient document retrieval. Within our framework, the entire retrieval process is encapsulated as a modular MCP tool, which the LLM can invoke autonomously based on its reasoning needs. 

Specifically, the Qwen3 model follows a structured interaction protocol in which reasoning and planning are carried out between \texttt{<think>} and \texttt{</think>} tags,tool invocation commands are embedded within \texttt{<tool\_call>} and \texttt{</tool\_call>}, and the corresponding outputs are captured between \texttt{<tool\_response>} and \texttt{</tool\_response>}. 
The model engages in multiple rounds of tool interaction until it either generates a final answer enclosed in \texttt{<answer></answer>} or reaches the maximum limit of tool invocations, which is set to six in our experiments. A concrete case study is presented in the following table to illustrate how this structured retrieval mechanism enables the model to access external knowledge and perform informed reasoning. 

The table ~\ref{prompt:system-user}presents a representative prompt designed to encourage tool usage, along with a concrete case study  ~\ref{case:tool-intergrated-qa}demonstrating how the model leverages structured tool interactions to retrieve external knowledge and perform reasoning.

\begin{tcolorbox}[
    float,
    floatplacement=htbp,
    colback=gray!5,
    colframe=gray!75,
    width=\textwidth,
    arc=1mm,
    boxrule=0.5pt,
    title=\textbf{Case 1: Example Response with Tool-Integrated Reasoning}
    halign title=center,
    label={case:tool-intergrated-qa}
]

response:\texttt{<think>}
Okay, let's tackle this question. The user is asking which U.S. state was historically dominated by the Republican party from 1860 to 1932 and is represented by State Senator George D. Maziarz.

First, I need to figure out the state in question. The key points here are the time period (1860-1932) and the Republican dominance. I remember that during the 19th century, the Republican Party was a major political force, especially after the Civil War. States like Ohio, New York, and Pennsylvania were known for their Republican strongholds. However, I'm not entirely sure which one fits the specific time frame mentioned.

Next, the mention of State Senator George D. Maziarz. I need to check if this senator is from a state that fits the historical Republican dominance. But I don't recall specific information about George D. Maziarz. Maybe I should look up his state of origin. However, since I can't access external resources right now, I need to rely on my existing knowledge.

...
\texttt{</think>}
\begin{verbatim}
<tool_call>
{"name": "search-query_rag", 
"arguments": {"query": "George D. Maziarz state", "topk": 3}}
</tool_call>}
\end{verbatim}
\texttt{<tool\_response>}
Execute the tool search-query\_rag successed
  - The args are: {'query': 'George D. Maziarz state', 'topk': 3}
  - The result is:......
\texttt{</tool\_response>}
\begin{verbatim}
<answer>New York</answer>
\end{verbatim}
\end{tcolorbox}

\subsection{Experimental Settings}\label{set}

\textbf{Direct Inference Baselines.}  
We compare our method against the following direct inference baselines:\textbf{Direct Inference + Chain-of-Thought (CoT)}~\citep{wei2022chain} performs reasoning solely through its internal chain of thought. \textbf{TIR (Tool-Integrated Reasoning)}~\citep{chai2025rlfactoryplugandplayreinforcementlearning} enhances inference by retrieving documents using integrated search tools. For \textbf{Reasoning with Memory}, we compare with three variants: (1) \textbf{Direct Trajectory}, which uses past trajectories directly as memory; (2) \textbf{A-MEM}~\citep{xu2025amemagenticmemoryllm}, which maintains a dynamic memory graph; and (3) \textbf{Expel}~\citep{zhao2024expelllmagentsexperiential}, which extracts high-level strategic insights from past experiences.

\textbf{Reinforcement Learning Baselines.}  
For RL training, we evaluate two groups of baselines:   \textbf{Search-R1:}~\citep{jin2025searchr1trainingllmsreason} A reinforcement learning agent that relies solely on multi-turn tool invocation without any memory support.  \textbf{RL with Memory Variants:} We examine the performance of agents equipped with the three memory types described above—Direct Trajectory, A-MEM, and Expel Memory—to assess how different memory designs impact training efficiency and overall performance.

We conduct experiments with two model scales, \textbf{Qwen-3-4B} and \textbf{Qwen-3-8B}. For the retrieval component, we adopt the 2018 Wikipedia dump~\citep{karpukhin2020densepassageretrievalopendomain} as the knowledge source and employ the \textbf{E5}~\citep{wang2024textembeddingsweaklysupervisedcontrastive} retriever.

We exclusively use the \textbf{HotpotQA} dataset, both for model optimization and for constructing memory during the memory formation process. Evaluation is then carried out on the test or validation sets of seven diverse datasets, enabling assessment of performance both within the training domain and in out-of-domain settings.  
We report \textbf{Exact Match (EM)} as the primary evaluation metric. And for memory construction in A-Mem~\citep{xu2025amemagenticmemoryllm}, Expel~\citep{zhao2024expelllmagentsexperiential}, and our proposed method, where a high-capability large language model is required, we utilized GPT-4o.

We conduct experiments on seven datasets, where HotpotQA is selected as the in-domain test set, while the remaining six datasets are used for out-of-domain evaluation. From the HotpotQA training set, we sample 1,000 examples to construct the memory and an additional 5,000 examples for weight training.

During the RL training phase, we use the rest of the HotpotQA training set as the training corpus. We adopt a batch size of 512 with a micro batch size of 64, and the rollout sampling is performed with a temperature of 1.0. To accelerate the rollout process of the LLM, we deploy vLLM v1 with a tensor parallel size of 1.

Specifically for the GRPO algorithm, the number of rollout samples (n) is set to 8. All experiments are conducted on a cluster of 8 NVIDIA A100 GPUs.

\section{Memory Graph Construction}
The pseudo-code for the overall process of the graph, which is composed of the specific paths of LLM models, is as shown in the algorithm~\ref{alg:memory_graph};

\subsection{The construction of Path in Finite State Machine}
\label{app:fsm}

\begin{sidewaysfigure}
    \centering
    \includegraphics[width=\textwidth]{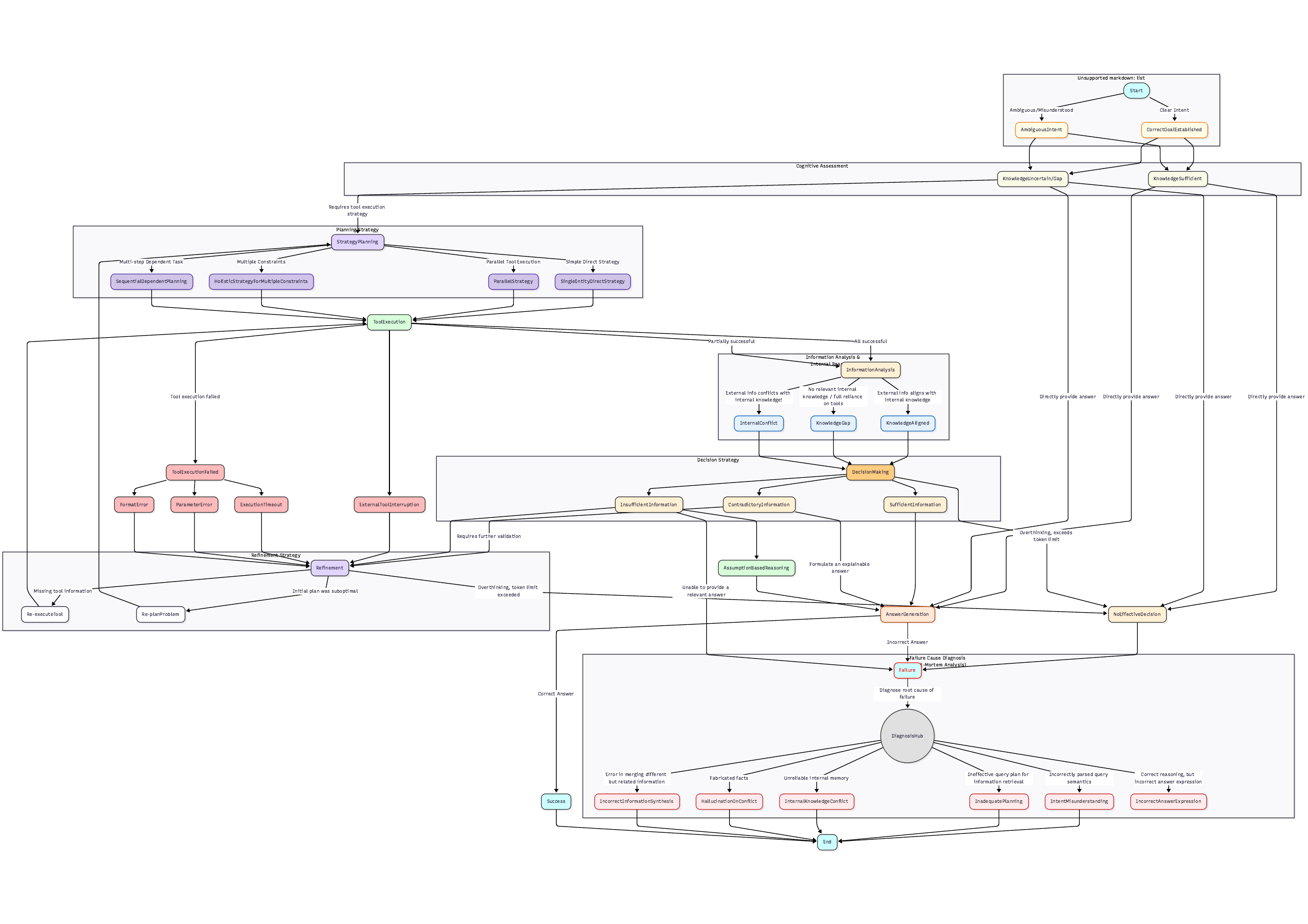}
    \caption{Finite State Machine}
    \label{fig:fsm}
\end{sidewaysfigure}

First, to formalize the agent's decision-making process during tool invocation, we construct a Finite State Machine , the overall architecture of which is depicted in Figure ~\ref{fig:fsm}. The states within this FSM are designed to encapsulate the critical cognitive junctures an LLM agent encounters, representing a synthesis of its internal knowledge and available external information. This design serves as a generalized abstraction of the agent's decision pathway, ensuring high generalizability across diverse tasks.

\begin{algorithm}[htbp]
\caption{Hierarchical Memory Graph Construction and Update}
\label{alg:memory_graph}
\begin{algorithmic}[1]
\State \textbf{Input:} Memory Graph $\mathcal{G}$, new query $q_i$, policy $\pi$, FSM $\mathcal{S}$, sample count $N$, similarity threshold $K$.
\State \textbf{Ensure:} Updated Memory Graph $\mathcal{G}'$.

\Procedure{UpdateMemoryGraph}{$\mathcal{G}, q_i, \pi, \mathcal{S}, N, K$}
    \State $T_s \leftarrow \emptyset, T_f \leftarrow \emptyset$ \Comment{Initialize sets for successful and failed paths}
    \State $\mathcal{G} \leftarrow \text{AddNode}(\mathcal{G}, q_i)$ \Comment{Add current query to the graph}

    \For{$n = 1$ \textbf{to} $N$} \Comment{Sample N trajectories from the policy}
        \State $\tau_n \leftarrow \text{SampleRollout}(\pi, q_i)$
        \State $t_n \leftarrow \text{GroundTrajectoryToPath}(\tau_n, \mathcal{S})$ \Comment{Map trajectory to a canonical FSM path}
        \State $\mathcal{G} \leftarrow \text{AddNode}(\mathcal{G}, t_n)$
        \State $\mathcal{G} \leftarrow \text{AddEdge}(\mathcal{G}, q_i, t_n)$
        
        \If{$\text{IsSuccess}(\tau_n)$}
            \State $T_s \leftarrow T_s \cup \{t_n\}$
        \Else
            \State $T_f \leftarrow T_f \cup \{t_n\}$
        \EndIf
    \EndFor
    
    \State $M_{new} \leftarrow \text{InduceMetaCognition}(q_i, T_s, T_f, \mathcal{G}, K)$ \Comment{Derive new strategic principles}
    
    \For{\textbf{each} new meta-cognition $m$ \textbf{in} $M_{new}$}
        \State $m_{exist} \leftarrow \text{FindMatchingMetaCognition}(m, \mathcal{G})$
        \If{$m_{exist}$ is \textbf{null}}
            \State $m_{final} \leftarrow \text{CreateNewMetaCognitionNode}(m)$
            \State $\mathcal{G} \leftarrow \text{AddNode}(\mathcal{G}, m_{final})$
        \Else
            \State $\text{UpdateConfidence}(m_{exist})$
            \State $m_{final} \leftarrow m_{exist}$
        \EndIf
        
        \For{\textbf{each} path $t$ that generated $m$} \Comment{Link paths to the principles they support}
             \State $\mathcal{G} \leftarrow \text{AddEdge}(\mathcal{G}, t, m_{final})$
        \EndFor
    \EndFor
    \State \Return $\mathcal{G}$
\EndProcedure
\vspace{1em}

\Procedure{InduceMetaCognition}{$q_i, T_s, T_f, \mathcal{G}, K$}
    \If{$T_s \neq \emptyset$ \textbf{and} $T_f \neq \emptyset$} \Comment{\textbf{Case 1:} High-confidence induction}
        \State $t_s \leftarrow \text{SelectOne}(T_s)$, $t_f \leftarrow \text{SelectOne}(T_f)$
        \State $m \leftarrow \text{ContrastPaths}(t_s, t_f)$ \Comment{e.g., find first diverging decision}
        \State \Return $\{m\}$
        
    \ElsIf{$T_s = \emptyset$ \textbf{and} $T_f \neq \emptyset$} \Comment{\textbf{Case 2:} Speculative induction}
        \State $M_{spec} \leftarrow \emptyset$
        \State $Q_{sim} \leftarrow \text{FindSimilarQueries}(q_i, \mathcal{G}, K)$ \Comment{Based on embedding similarity}
        \For{\textbf{each} similar query $q_j$ \textbf{in} $Q_{sim}$}
            \For{\textbf{each} successful path $t_j$ of $q_j$}
                \State $M_{spec} \leftarrow M_{spec} \cup \text{GetMetaCognitionsFromPath}(t_j, \mathcal{G})$
            \EndFor
        \EndFor
        \State \Return $M_{spec}$
    \Else
        \State \Return $\emptyset$ \Comment{No new insights if only successes or no rollouts}
    \EndIf
\EndProcedure
\end{algorithmic}
\end{algorithm}

We first have a Qwen3 series model produce a concrete answer to the query. Subsequently, to map an agent's raw execution trajectory—the specific sequence of reasoning and tool calls—onto a canonical path within the FSM, we leverage a powerful large language model, see Table ~\ref{prompt:fsm-construction} for the full prompt and the table ~\ref{case:fsm}presents a specific case of Finite State Machine.

\begin{tcolorbox}[
    float,
    floatplacement=htbp,
    colback=gray!5,
    colframe=gray!75,
    width=\textwidth,
    arc=1mm,
    boxrule=0.5pt,
    title=\textbf{Case 2: A example of Finite State Machine},
    halign title=center,
    label={case:fsm}
]
\paragraph{Illustrative Decision Path.}
The following sequence illustrates a canonical decision path encoded within our framework:
\[
\begin{aligned}
\text{Start} &\rightarrow \text{CorrectGoalEstablished} \rightarrow \text{KnowledgeUncertainGap} \\
&\rightarrow \text{StrategyPlanning} \rightarrow \text{SequentialDependentPlanning} \\
&\rightarrow \text{ToolExecution} \rightarrow \text{InformationAnalysis} \\
&\rightarrow \text{KnowledgeAligned} \rightarrow \text{DecisionMaking} \\
&\rightarrow \text{InsufficientInformation} \rightarrow \text{AssumptionBasedReasoning} \\
&\rightarrow \text{AnswerGeneration} \rightarrow \text{WrongAnswer} \\
&\rightarrow \text{DiagnosisHub} \rightarrow \text{InternalKnowledgeConflict} \rightarrow \text{End}
\end{aligned}
\]
This path represents a chain of cognitive states traversed by the agent. It begins with goal establishment, proceeds through planning and execution, encounters a knowledge gap leading to flawed reasoning, and concludes with self-diagnosis. By encoding such trajectories as nodes in the transition path layer, the graph provides a structured and abstract representation of a complex reasoning process, which can be analyzed, compared, and learned from.
\end{tcolorbox}

\subsection{Meta-cognition Construction}
The detailed descriptions of each type of node in the memory are as follows:
\begin{itemize}
    \item \textbf{Query Layer} $\mathcal{Q}$: Each node $q_i \in \mathcal{Q}$ represents a specific task instance, such as a user-issued query. It encapsulates the entirety of an interaction, including the initial input, the agent's generated output, the complete execution trajectory, and a resultant outcome label (e.g., success or failure).
    
    \item \textbf{Transition Path Layer} $\mathcal{T}$: Each node $t_j \in \mathcal{T}$ corresponds to a standardized decision-making pathway. These pathways are grounded in a predefined finite state machine (FSM) $\mathcal{S}$, representing a canonical sequence of the agent's states and actions. This layer abstracts away instance-specific details to reveal underlying behavioral patterns.
    
    \item \textbf{Meta-Cognition Layer} $\mathcal{M}$: Each node $m_k \in \mathcal{M}$ encodes a high-level, human-readable strategic principle. These principles are distilled from a comparative analysis of successful and failed transition paths, representing generalized heuristics for effective problem-solving.
\end{itemize}

The induction of meta-cognitions is accomplished through three primary analytical scenarios, each facilitated by a dedicated prompt:

Intra-Query Analysis: This involves comparing successful and failed trajectories that originate from the identical query to distill a high-confidence causal principle. The prompt for this process is presented in ~\ref{prompt:meta-construction}.

Inter-Query Analysis: This contrasts a failed trajectory with successful ones from semantically similar but distinct queries to generate speculative heuristics. 

Positive Example Distillation: This process extracts generalizable strategies exclusively from a collection of successful execution paths. 

\section{Experiment analysis}

\subsection{the number of the meta-cognition}
To better understand how the quantity of retrieved meta-cognitive strategies affects agent performance, we evaluate four configurations: using 0 (no memory, denoted as ITR), 1, 3, and 5 strategies as contextual input. Results across seven QA benchmarks are presented in Table~\ref{tab:meta-number-ablation}.

We observe that introducing even a single meta-cognitive strategy leads to a notable improvement over the baseline (ITR), especially on multi-hop tasks such as Bamboogle (+11.6\%) and HotpotQA (+2.9\%). This suggests that explicit strategic signals can substantially aid reasoning even in limited quantities. As the number of strategies increases, performance generally improves, but the marginal gains become smaller—likely due to redundancy or prompt saturation. The best overall result is achieved at \texttt{top\_k=5}, which balances diversity and relevance.

These findings imply that a moderate number of well-curated strategies can enhance generalization and decision quality, without incurring the risks of prompt overload or noise from irrelevant memories.

\begin{table*}[htpb]
\centering
\caption{Performance of different numbers of meta-cognition.}
\label{tab:meta-number-ablation}
\resizebox{\textwidth}{!}{
\begin{tabular}{l|ccc|ccccc|c}
\hline
\multirow{2}{*}{Methods} & \multicolumn{3}{c|}{General QA} & \multicolumn{5}{c|}{Multi-Hop QA} & \multirow{2}{*}{Avg.} \\ \cline{2-9}
 & NQ$^\star$ & TriviaQA$^\star$ & PopQA$^\star$ & HotpotQA$^\dagger$ & 2wiki$^\star$ & Musique$^\star$ & Bamboogle$^\star$ &  \\ \hline
\multicolumn{10}{l}{\textbf{Qwen3-4B}} \\
ITR              &    0.298   &    0.581   &   0.157    &   0.268    &   0.281    &  0.077     &   0.290    &   &  0.279  \\
$top_k=1$       &    0.326   &   0.583    &   0.382    &    0.290   &  0.327     &   0.096    &   0.406    &   &0.344 \\
$top_k=3$       &   0.335    &   0.596    &  0.393    &   0.299    &    0.347   &   0.099    &   0.391   &  &  0.351   \\
$top_k=5$         &   0.333    &  0.594     &  0.392      &   0.299    &   0.349    &    0.094   &    0.418   &   &   0.355  \\ \hline

\end{tabular}
}
\end{table*}

\subsection{Cross-API Memory Robustness}
To further validate the portability and reliability of our structured memory graph, we construct the memory using two distinct LLM APIs: \texttt{gpt-4o} and \texttt{Gemini-2.5-pro}. These memory graphs are then integrated into the same downstream agent architecture (Qwen3-4B and Qwen3-8B), and evaluated across seven QA datasets. As shown in Table~\ref{tab:diff-api}, the resulting performance differences are minor, with Gemini-based memory slightly outperforming its 4o counterpart in most cases.

Specifically, on the multi-hop benchmark Bamboogle, the Gemini-constructed memory shows a notable increase (e.g., +0.043 on Qwen3-8B), while maintaining parity or marginal gains in general QA datasets like TriviaQA and PopQA. These results indicate that while different APIs may introduce slight variations in strategy abstraction, our framework is robust to such differences and maintains high effectiveness regardless of the underlying model used to generate the memory.

\begin{table*}[htbp]
\centering
\caption{Performance comparison across LLM .}
\label{tab:diff-api}
\resizebox{\textwidth}{!}{
\begin{tabular}{l|ccc|ccccc|c}
\hline
\multirow{2}{*}{Methods} & \multicolumn{3}{c|}{General QA} & \multicolumn{5}{c|}{Multi-Hop QA} & \multirow{2}{*}{Avg.} \\ \cline{2-9}
 & NQ$^\star$ & TriviaQA$^\star$ & PopQA$^\star$ & HotpotQA$^\dagger$ & 2wiki$^\star$ & Musique$^\star$ & Bamboogle$^\star$ &  \\ \hline
\multicolumn{10}{l}{\textbf{Qwen3-8B}} \\
Ours(4o)        &    0.316   &   0.622    &   0.382    &   0.358    &   0.354    &   0.128    &   0.392    & &  0.365 \\
Ours(gemini)           &   0.318    &   0.621    &   0.385    &   0.362    &   0.336    &   0.123    &    0.434  &  &   0.369  \\ \hline

\multicolumn{10}{l}{\textbf{Qwen3-4B}} \\
Ours(4o)        &   0.335    &   0.596    &  0.393    &   0.299    &    0.347   &   0.099    &   0.391   &  &  0.351   \\
Ours(gemini)           &   0.337    &    0.598   &  0.396   &    0.314   &    0.360   &   0.093    & 0.397   & &    0.357   \\ \hline

\end{tabular}
}
\end{table*}

\begin{algorithm}[htbp]
\caption{Trainable Graph Weight Optimization via Policy Gradient}
\label{alg:weight_optimization}
\begin{algorithmic}[1]
\State \textbf{Input:} Memory Graph $\mathcal{G}$ with initial weights $\mathbf{w}$, Training Queries $\mathcal{D}$, Agent model, Reward function $\mathcal{R}$, learning rate $\alpha$.
\State \textbf{Output:} Optimized Memory Graph $\mathcal{G}$ with updated weights $\mathbf{w}^*$.

\Procedure{OptimizeGraphWeights}{$\mathcal{G}, \mathcal{D}, \alpha$}
    \For{\textbf{each} query $q_{\text{new}}$ \textbf{in} $\mathcal{D}$}
        \State \Comment{--- Step 1: Stochastic Guidance Selection ---}
        \State $m_k, p(m_k \mid q_{\text{new}}) \leftarrow \text{SelectGuidingMetaCognition}(\mathcal{G}, q_{\text{new}})$
        
        \If{$m_k$ is \textbf{null}} \Comment{No relevant guidance found}
            \State \textbf{continue}
        \EndIf
        
        \State \Comment{--- Step 2: Counterfactual Evaluation ---}
        \State Response$_{\text{with}} \leftarrow \text{Agent.generate}(q_{\text{new}}, \text{guidance}=m_k)$
        \State $R_{\text{with}} \leftarrow \mathcal{R}(\text{Response}_{\text{with}}, q_{\text{new}})$
        
        \State Response$_{\text{w/o}} \leftarrow \text{Agent.generate}(q_{\text{new}}, \text{guidance}=\text{null})$
        \State $R_{\text{w/o}} \leftarrow \mathcal{R}(\text{Response}_{\text{w/o}}, q_{\text{new}})$
        
        \State $\Delta R_k \leftarrow R_{\text{with}} - R_{\text{w/o}}$ \Comment{Calculate reward gap (utility signal)}
        
        \State \Comment{--- Step 3: Policy Gradient Update ---}
        \State $\nabla_{\mathbf{w}} \mathcal{L} \leftarrow - \Delta R_k \cdot \nabla_{\mathbf{w}} \log p(m_k \mid q_{\text{new}})$ \Comment{Compute gradient for REINFORCE}
        \State $\mathbf{w} \leftarrow \mathbf{w} - \alpha \cdot \nabla_{\mathbf{w}} \mathcal{L}$ \Comment{Update all contributing weights}
    \EndFor
    \State \Return $\mathcal{G}$
\EndProcedure
\vspace{1em}

\Procedure{SelectGuidingMetaCognition}{$\mathcal{G}, q_{\text{new}}$}
    \State \Comment{Activate relevant subgraph based on semantic similarity}
    \State $\mathcal{M}_{\text{act}} \leftarrow \text{ActivateSubgraph}(q_{\text{new}}, \mathcal{G})$
    
    \If{$\mathcal{M}_{\text{act}}$ is empty}
        \State \Return \textbf{null}, 0
    \EndIf

    \State \Comment{Compute relevance scores for all activated meta-cognitions}
    \ForAll{$m \in \mathcal{M}_{\text{act}}$}
        \State $S(m \mid q_{\text{new}}) \leftarrow 0$
        \ForAll{path $q_i \rightarrow t_j \rightarrow m$ in $\mathcal{G}$}
            \If{$q_i$ is in activated subgraph}
                \State $S(m \mid q_{\text{new}}) \leftarrow S(m \mid q_{\text{new}}) + \text{Sim}(q_{\text{new}}, q_i) \cdot w_{qt}^{(i,j)} \cdot w_{tm}^{(j,m)}$
            \EndIf
        \EndFor
    \EndFor
    
    \State \Comment{Compute selection probabilities using softmax}
    \State $Z \leftarrow \sum_{m' \in \mathcal{M}_{\text{act}}} \exp(S(m' \mid q_{\text{new}}))$
    \ForAll{$m \in \mathcal{M}_{\text{act}}$}
        \State $p(m \mid q_{\text{new}}) \leftarrow \exp(S(m \mid q_{\text{new}})) / Z$
    \EndFor
    
    \State \Comment{Stochastically sample a meta-cognition based on probabilities}
    \State $m_k \leftarrow \text{Sample}(\mathcal{M}_{\text{act}}, \text{probabilities}=\{p(m \mid q_{\text{new}})\})$
    
    \State \Return $m_k, p(m_k \mid q_{\text{new}})$
\EndProcedure
\end{algorithmic}
\end{algorithm}

\section{Case Studies}

This case ~\ref{case:meta-cog-correction}illustrates how the integration of meta-cognitive strategies enhances factual precision. Without meta-cognition, the agent returns a partially correct but under-specified answer ("National Security Law"). With meta-cognition, the agent engages in structured validation and corrects the response to the fully grounded and jurisdiction-specific "Macau National Security Law", aligning exactly with the ground truth. This aligns with our design goal: equipping agents with self-checking mechanisms that reduce synthesis ambiguity, particularly when internal confidence is high but error risk remains.

\begin{tcolorbox}[
    float,
    floatplacement=htbp,
    colback=gray!5,
    colframe=gray!75,
    width=\textwidth,
    arc=1mm,
    boxrule=0.5pt,
    title=\textbf{Case 3: Correcting Answer via Meta-Cognition},
    halign title=left,
    fonttitle=\bfseries,
    label={case:meta-cog-correction}
]
\textbf{Query:} \textit{Hong Kong Macau cultural exchange was a trip that tested which law whose purpose was to fulfil Article 23 of the Macau Basic Law?}

\vspace{1mm}
\textbf{Ground Truth:} Macau National Security Law

\vspace{1mm}
\textbf{Meta-Cognition:} \textit{Early recognition and affirmation of the 'KnowledgeSufficient' state can help prevent synthesis inaccuracies in scenarios where internal confidence is prone to challenges. Structured validation strategies ensure precise answer formulation when internal knowledge seems sufficient but risks synthesis errors.}

\vspace{1mm}
\textbf{Response (w/o meta-cognition):}
\begin{quote}
\textit{... Therefore, the answer should state that the law tested was the National Security Law, aimed at fulfilling Article 23.}  
\textbf{→ Final Answer:} \texttt{National Security Law}
\end{quote}

\vspace{1mm}
\textbf{Response (with meta-cognition):}
\begin{quote}
\textit{... That law is likely the Macau National Security Law, which was enacted in 2009... The answer should be the Macau National Security Law.}  
\textbf{→ Final Answer:} \texttt{Macau National Security Law}
\end{quote}

\end{tcolorbox}

\section{prompt templates}

\subsection{multi-turn tool-intergrated QA prompt}
When LLMs needs to interact with tools multiple times to answer a question, it is necessary to guide the LLM on which tools to use and how to use them. The specific prompt is as follows.
\begin{tcolorbox}[
    float,
    floatplacement=htbp,
    colback=gray!5,
    colframe=gray!75,
    width=\textwidth,
    arc=1mm,
    boxrule=0.5pt,
    title=\textbf{Prompt A: System and User Prompt},
    halign title=center,
    label={prompt:system-user}
]
\textbf{SYSTEM PROMPT:}

\begin{lstlisting}[basicstyle=\ttfamily\footnotesize, breaklines=true]
# Tools
You may call one or more functions to assist with the user query.
You are provided with function signatures within <tools></tools> XML tags:
<tools>
{
  "name": "search-query_rag",
  "description": "MCP RAG Query Tool (Synchronous Version)
  Args:
    query: query text
    topk: The default number of documents returned is 3
  Returns:
    str: The formatted query result",
  "parameters": {
    "type": "object",
    "properties": {
      "query": {"title": "Query", "type": "string"},
      "topk": {"default": 3, "title": "Topk", "type": "integer"}
    },
    "required": ["query"]
  }
}
</tools>

# Tool call format
For each function call, return a JSON object with function name and arguments within <tool_call></tool_call> XML tags:
<tool_call>
{
  "name": <function-name>,
  "arguments": <args-json-object>
}
</tool_call>
\end{lstlisting}

\vspace{0.8em}
\textbf{USER PROMPT:}

\begin{lstlisting}[basicstyle=\ttfamily\footnotesize, breaklines=true]
Answer the given question. After reasoning, if you find you lack some knowledge, you can call the search tool.
You may search as many times as you want.
If you find no further external knowledge is needed, you can directly provide the answer inside <answer> and </answer>, without detailed illustrations.
For example: <answer> Beijing </answer>.

Question: Which US State, historically dominated by the Republican party from 1860 to 1932, is represented by State Senator George D. Maziarz?
\end{lstlisting}

\end{tcolorbox}

\subsection{constructing fsm path}
Given the specific response path of the LLM and the complete structure of the state machine, we employ an LLM (e.g., GPT-4o) to map the generated answer onto one of the predefined paths in the state machine. The following shows the exact prompt used.
\begin{tcolorbox}[
    float,
    floatplacement=htbp,
    colback=gray!5,
    colframe=gray!75,
    width=\textwidth,
    arc=1mm,
    boxrule=0.5pt,
    title=\textbf{Prompt B: Prompt for constructing FSM path},
    halign title=center,
    label={prompt:fsm-construction}
]
\textbf{Instruction:} You are a metacognition analysis expert specialized in extracting \textit{generalized decision principles and guidance strategies} from state machine execution paths.

\textbf{State machine transition rules:} \texttt{\{transitions\_info\}}

\textbf{Core Requirements:}
\begin{enumerate}[leftmargin=2em]
    \item \textbf{Generalizability Focus}: Output strategies and principles must be general, applicable to similar problems, without specific query details.
    \item \textbf{Direct Usability}: Generated content should be directly usable as guidance principles for new problems.
    \item \textbf{Principled Expression}: Use cautious guidance terms like ``consider'', ``may help'', ``tends to'' rather than definitive statements.
    \item \textbf{Concise Effectiveness}: Output only the most core insights, avoid redundancy and complexity.
    \item \textbf{Quality Control}: Strictly evaluate whether there is sufficient evidence to support new metacognition.
    \item \textbf{Knowledge Confidence Awareness}: Recognize that LLM's internal knowledge confidence varies across queries --- success patterns may be domain-specific.
    \item \textbf{Uncertainty Acknowledgment}: Express appropriate uncertainty in guidance principles, avoiding overly definitive conclusions.
    \item \textbf{Quantity Management}: When metacognition count exceeds 30, prioritize updating low-confidence existing metacognitions.
\end{enumerate}

\vspace{0.8em}
\textbf{Output Format (Quantity-Aware):}

Your output must be a JSON object with the following structure:

\begin{lstlisting}[basicstyle=\ttfamily\footnotesize, breaklines=true]
{
  "decision": "update" or "create" or "skip",
  "target_meta_id": <ID of metacognition to update (only when decision is "update")>,
  "reasoning": "Brief explanation including quantity management when count > 30.",
  "meta_cognition": {
    "summary": "Concise general guidance summary (use cautious language).",
    "strategy_principles": [
      {
        "principle": "...",
        "confidence": "high" | "medium" | "low",
        "confidence_score": 30 - 85
      },
      ...
    ],
    "overall_confidence": "high" | "medium" | "low",
    "evidence_paths": <int>,
    "uncertainty_note": "Brief acknowledgment of limitations or knowledge-dependency concerns."
  }
}
\end{lstlisting}

\end{tcolorbox}

\subsection{meta-cognition constructing}
With both successful and failed state-machine paths available, we derive high-level meta-cognitions by contrasting the two. The following prompt illustrates how a pair of successful and failed paths under the same query is used to induce meta-cognition.
\begin{tcolorbox}[
    float,
    floatplacement=htbp,
    colback=gray!5,
    colframe=gray!40!black,
    width=\textwidth,
    arc=1mm,
    boxrule=0.5pt,
    title=\textbf{Prompt C: Metacognition Prompt Specification},
    halign title=left,
    breakable,
    label={prompt:meta-construction}
]

\textbf{State machine transition rules:} \texttt{\{transitions\_info\}}

\vspace{0.8em}
\textbf{Core Requirements:}
\begin{enumerate}[leftmargin=2em]
    \item \textbf{Generalizability Focus:} Output strategies and principles must be general, applicable to similar problems, without specific query details.
    \item \textbf{Direct Usability:} Generated content should be directly usable as guidance principles for new problems.
    \item \textbf{Principled Expression:} Use cautious guidance terms like ``consider'', ``may help'', ``tends to'' rather than definitive statements.
    \item \textbf{Concise Effectiveness:} Output only the most core insights, avoid redundancy and complexity.
    \item \textbf{Quality Control:} Strictly evaluate whether there is sufficient evidence to support new metacognition.
    \item \textbf{Knowledge Confidence Awareness:} \textbf{Recognize that LLM's internal knowledge confidence varies across queries---success patterns may be domain-specific.}
    \item \textbf{Uncertainty Acknowledgment:} \textbf{Express appropriate uncertainty in guidance principles, avoiding overly definitive conclusions.}
    \item \textbf{Quantity Management:} When metacognition count exceeds 30, prioritize updating low-confidence existing metacognitions.
\end{enumerate}

\vspace{0.5em}
\textbf{Critical Self-Reflection Requirements:}
\begin{itemize}[leftmargin=2em]
    \item \textbf{Pattern Validity:} Question whether identified patterns truly represent generalizable principles.
    \item \textbf{Knowledge Dependency:} \textbf{Consider if success stems from strategy effectiveness or the LLM's domain familiarity.}
    \item \textbf{Evidence Sufficiency:} Demand higher evidence standards for strategies that could mislead future queries.
    \item \textbf{Simplicity Over Complexity:} Favor simple, robust principles over complex, brittle ones.
\end{itemize}

\vspace{0.5em}
\textbf{Metacognition Quantity Control Strategy:}
\begin{description}[leftmargin=2em, style=nextline]
\item[When metacognition count $\leq$ 30:] 
  \begin{itemize}
    \item Normal decision making: create, update, or skip based on evidence quality.
    \item Prefer creating new metacognition when patterns are sufficiently distinct.
    \item \textbf{Express appropriate uncertainty in new metacognitions.}
  \end{itemize}

\item[When metacognition count $>$ 30:]
  \begin{itemize}
    \item \textbf{Strongly prefer UPDATE over CREATE}: Prioritize improving existing low-confidence metacognitions.
    \item Only create new metacognition if the pattern is exceptionally valuable and completely distinct.
    \item Target metacognitions with confidence levels ``low'' or ``medium'' for updates.
  \end{itemize}
\end{description}

\vspace{0.5em}
\textbf{Analysis Focus:}
\begin{enumerate}[leftmargin=2em]
    \item \textbf{Success Pattern Identification:} Abstract reusable decision patterns from successful paths.
    \item \textbf{Failure Cause Summary:} Identify generalizable errors to avoid from failed paths.
    \item \textbf{State Transition Optimization:} Extract best practice principles for state machine execution.
    \item \textbf{Knowledge Dependency Assessment:} \textbf{Evaluate whether patterns might be specific to certain knowledge domains.}
    \item \textbf{Existing Knowledge Enhancement:} When quantity is high, focus on strengthening weak metacognitions.
\end{enumerate}

\vspace{0.5em}
\textbf{Decision Options:}
\begin{itemize}[leftmargin=2em]
    \item \textbf{create:} Create new metacognition (when discovering valuable and distinct patterns, or when quantity $\leq$ 30).
    \item \textbf{update:} Update existing metacognition (preferred when quantity $>$ 30, especially targeting low-confidence ones).
    \item \textbf{skip:} Skip metacognition operation (when evidence is insufficient or has no new value).
\end{itemize}

\vspace{0.5em}
\textbf{Skip Metacognition Situations:}
\begin{itemize}[leftmargin=2em]
    \item Path data quality is poor, patterns are unclear.
    \item Existing metacognition already covers the pattern, new evidence shows no significant improvement.
    \item Success/failure path differences are not obvious, difficult to extract effective strategies.
    \item \textbf{Cannot distinguish whether success stems from strategy effectiveness or knowledge domain familiarity.}
    \item When quantity $>$ 30 and no suitable low-confidence metacognition found for update.
\end{itemize}

\vspace{0.5em}
\textbf{Output Format (Quantity-Aware):}
Your output must be a JSON object containing:

\begin{lstlisting}[basicstyle=\ttfamily\footnotesize, breaklines=true]
{
  "decision": "update" or "create" or "skip",
  "target_meta_id": (when decision is update) ID of metacognition to update,
  "reasoning": "Brief decision analysis, must include quantity management when count > 30.",
  "meta_cognition": {
    "summary": "...",
    "strategy_principles": [
      {"principle": "...", "confidence": "high", "confidence_score": 80},
      {"principle": "...", "confidence": "medium", "confidence_score": 60}
    ],
    "overall_confidence": "medium",
    "evidence_paths": 7,
    "uncertainty_note": "..." 
  }
}
\end{lstlisting}

\end{tcolorbox}
\end{document}